\documentclass[10pt,twocolumn,letterpaper]{article}

\usepackage{cvpr}              %

\usepackage[export]{adjustbox}
\definecolor{cvprblue}{rgb}{0.21,0.49,0.74}
\usepackage[pagebackref,breaklinks,colorlinks,allcolors=cvprblue]{hyperref}
\usepackage{graphicx}
\usepackage{tcolorbox}

\def\paperID{20} %
\def\confName{CVPR}
\def\confYear{2026}

\title{ConfusionBench: An Expert-Validated Benchmark for Confusion Recognition and Localization in Educational Videos }

\author{%
Lu Dong\thanks{These authors contributed equally.}\ , 
Xiao Wang\footnotemark[1] , 
 Mark Frank, Srirangaraj Setlur, Venu Govindaraju, Ifeoma Nwogu\\
State University of New York at Buffalo, Buffalo, NY 14260, USA\\
{\tt\small \{ludong, xwang277, mfrank83, setlur, govind, inwogu\}@buffalo.edu}
}

\makeatletter
\apptocmd{\@maketitle}{\myfigure}{}{}%
\makeatother

\begin{document}

\newcommand\myfigure{%
\centering
    \includegraphics[width=1\linewidth]{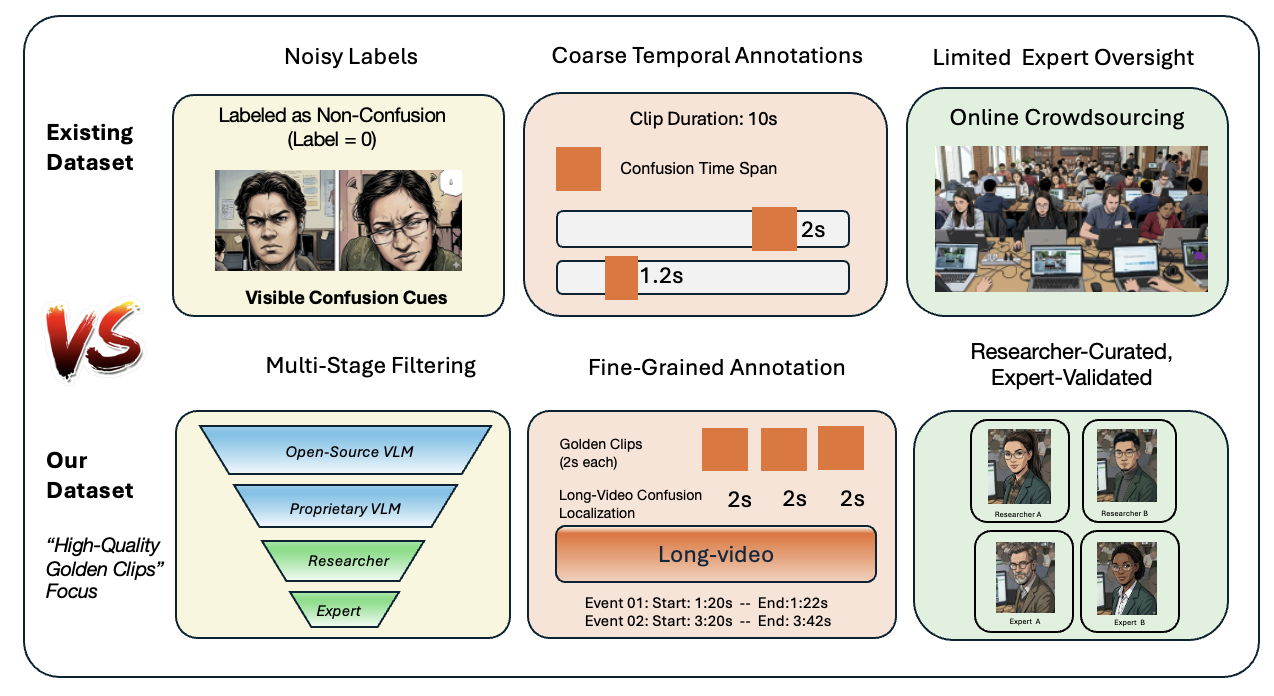}
\captionof{figure}{Comparison between existing confusion datasets and our ConfusionBench. Existing datasets often suffer from noisy labels, coarse temporal annotations, and limited expert oversight. In contrast, our ConfusionBench highlights multi-stage filtering, fine-grained annotation, and expert-validated golden clips, together with long-video confusion localization. All human figures shown in this illustration are AI-generated.
}
\label{fig:quality}
\vspace{0.5em}
}

\maketitle

\begin{abstract}
Recognizing and localizing student confusion from video is an important yet challenging problem in educational AI. Existing confusion datasets suffer from noisy labels, coarse temporal annotations, and limited expert validation, which hinder reliable fine-grained recognition and temporally grounded analysis. 
To address these limitations, we propose a practical multi-stage filtering pipeline that integrates two stages of model-assisted screening, researcher curation, and expert validation to build a higher-quality benchmark for confusion understanding.
Based on this pipeline, we introduce ConfusionBench, a new benchmark for educational videos consisting of a balanced confusion recognition dataset and a video localization dataset.
We further provide zero-shot baseline evaluations of a representative open-source model and a proprietary model on clip-level confusion recognition, long-video confusion localization tasks.
Experimental results show that the proprietary model performs better overall but tends to over-predict transitional segments, while the open-source model is more conservative and more prone to missed detections.
In addition, the proposed student confusion report visualization can support educational experts in making intervention decisions and adapting learning plans accordingly. 
All datasets and related materials will be made publicly available on our project page.

\end{abstract}

\section{Introduction}
\label{sec:intro}

Emotions are psycho bio social reactions that motivate actions \cite{matsumoto2012nonverbal}. These actions are reflected in the nonverbal behavior of people, primarily in the face \cite{frank2016evolution}. 
Although seven or eight emotions have received the most attention\cite{dong2024ig3d}, recent research suggests that scientists have identified up to 28 distinct emotions, including those associated with higher cognitive activity, such as realization, concentration, and confusion \cite{cowen2019mapping}.
Of these emotions, confusion turns out to be one of the most relevant to the education process \cite{myers2021automatic}. 
Identifying the specifics of that particular expression would be imperative to any teacher or automated system designed to assess the efficacy of teaching and learning. 
In today’s era of rapidly advancing artificial intelligence\cite{dong2024word,dong2024signavatar,zhai2023language,shan2024towards}, reliable recognition of student confusion and automatic visualization of confusion states can further support a wide range of educational AI applications, including intelligent tutoring systems, adaptive feedback, learner state monitoring, expert intervention analysis, and human-centered educational interfaces \cite{d2009responding,myers2021automatic,zeng2017learner,wang2025automisty,wang2026mistypilot}.
Reliable confusion recognition and automatic visualization of student confusion states can further support a broad range of educational AI applications, including intelligent tutoring systems, adaptive feedback, learner monitoring, expert intervention analysis, and human-centered educational interfaces \cite{d2009responding,myers2021automatic,zeng2017learner}.
Meanwhile, recent advances in multimodal foundation models have sparked growing interest in using vision-language models (VLMs) to interpret human affective and cognitive states\cite{yu2025exploring}. However, VLM-driven student confusion analysis remains underexplored. 

A key challenge is that existing confusion datasets often do not provide sufficiently reliable annotations for evaluation. 
DAISEE is a valuable and publicly available dataset with confusion annotations, and has served as an important resource for prior research \cite{gupta2016daisee}.
However, it has several limitations that hinder its use as a reliable benchmark for confusion recognition. 
First, the labels can be noisy: some clips annotated as non-confusion still exhibit visible cues of confusion, while some clips labeled as confusion contain little or no observable evidence of confusion \cite{khan2024revisiting}. 
Second, the temporal annotations are overly coarse. Assigning a single confusion label to a 10-second clip may be insufficient, as the actual confusion cue may appear for only 1–2 seconds. As a result, such clip-level annotations fail to accurately capture the video’s fine-grained temporal cognitive state \cite{mohamad2019automatic,komaravalli2023detecting}.
Third, the annotation process often depends on large-scale online crowdsourcing, which can lead to substantial subjective inconsistency in the absence of sufficient expert validation. 
As a result, the limitations of such datasets make it challenging to reliably evaluate model performance on confusion detection in realistic educational settings. Therefore, high-quality and reliable confusion annotations are essential for meaningful evaluation \cite{khan2024revisiting,gupta2016daisee}.

To address these issues, expert validation is essential for ensuring annotation quality. However, immersing experts in large volumes of raw data is both inefficient and costly. To this end, we propose an efficient model-assisted benchmark construction pipeline that ensures data quality while making effective use of limited expert resources.
The core idea of this pipeline is to construct short, high-confidence golden clips through a multi-stage filtering process that includes coarse screening with open-source VLMs, fine-grained screening with commercial VLMs, researcher curation, and expert validation.
These short clips provide a more reliable reference standard for confusion recognition, particularly for defining confusion states and identifying subtle expressions and motion combinations that are often diluted in longer temporal segments.

Moreover, since DAiSEE is originally segmented from longer source videos, the validated golden clips can be traced back to their original temporal positions. Based on these positions and expert annotations as anchors, we can efficiently propagate annotations to neighboring frames with similar visual patterns, thereby enabling timestamp-level labels for long-video confusion event detection. 
This long-video setting is especially valuable in educational applications, where the frequency and temporal distribution of confusion over time are often more informative than isolated clip-level labels.
Based on this pipeline, we present a benchmark consisting of two datasets. The first is a balanced confusion recognition dataset, in which all 2-second confusion samples are expert-validated, while non-confusion samples are selected to exclude any confusion patterns identified in prior literature.
The second is a confusion localization dataset, where each long video is annotated with the start and end timestamps of confusion events, enabling evaluation of zero-shot confusion localization performance. In addition, we propose an efficient visualization report to facilitate expert intervention analysis and adaptive learning support.

Beyond benchmark construction, another goal of this work is to study how well modern VLMs can recognize confusion. Although recent open-source and proprietary vision-language models have shown strong zero-shot performance on many visual reasoning tasks, their ability to detect subtle confusion signals in temporally localized educational settings remains unclear. To address this gap, we provide zero-shot baseline evaluations of representative open-source and proprietary models on both clip-level confusion recognition and long-video confusion localization. Our contributions are three-fold:

\begin{itemize}
    \item We identify several key limitations of existing confusion datasets, including noisy labels, overly coarse temporal annotations, and insufficient expert validation, and propose a practical multi-stage pipeline for constructing higher-quality confusion benchmarks.

    \item We introduce a high-quality, expert-validated benchmark for confusion understanding in educational videos, consisting of a balanced recognition dataset, long videos with fine-grained temporal annotations, and a confusion report visualization design to support interpretation, intervention analysis, and adaptive learning.

    \item We benchmark modern VLMs in the zero-shot setting on clip-level confusion recognition and long-video confusion localization, establishing systematic baselines for confusion understanding in educational videos.
\end{itemize}

\section{Related Work}
\label{sec:formatting}

\begin{figure}[t]
    \centering
     \includegraphics[width=1.0\linewidth]{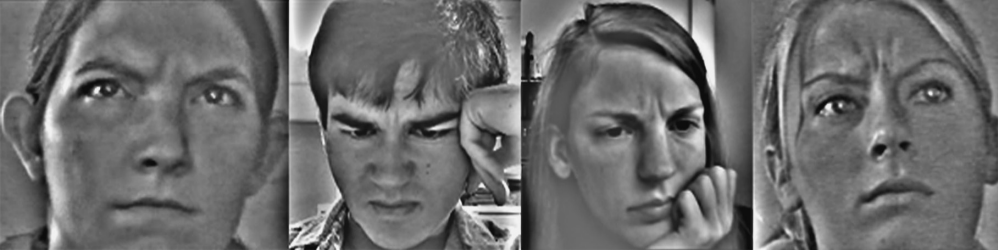}
    \caption{Facial Expressions of Confusion ~\cite{d2014confusion}.}
    \label{fig:confusion}
\end{figure}

\paragraph{Confusion Research}
Confusion plays a central role in complex learning activities, such as understanding difficult texts, generating coherent arguments, solving challenging problems, and modeling complex systems, and is often regarded as an inevitable consequence of effortful information processing~\cite{d2014confusion}. 
Prior research~\cite{d2014confusion,d2012dynamics,lehman2012confusion,grafsgaard2011predicting} suggests that confusion is associated with characteristic facial, bodily, and temporal cues. In particular, using the Facial Action Coding System (FACS)~\cite{prince2015facial}, D’Mello and colleagues~\cite{d2014confusion} identified several facial actions associated with confusion, including brow lowering or frowning (AU4), eyelid tightening or squinting (AU7), upper lip raising (AU10), and lip pressing or tightening (AU24). Among these, AU4, AU7, and especially their combination (AU4+AU7), have been reported as the most reliable facial correlates of confusion, as shown in Fig.~\ref{fig:confusion}. Additional contextual cues, such as gaze direction and head pose, may also provide supportive evidence in practice~\cite{rozin2003high}.

Beyond facial cues, researchers have also studied hand-to-face behaviors, body posture, and temporal dynamics as useful signals of confusion. Hand-to-face actions such as touching the chin, pressing the forehead, and covering the mouth may indicate thinking, frustration, or hesitation~\cite{mahmoud2011interpreting}, while body posture and movement patterns can provide complementary evidence~\cite{karg2013body,d2007posture,mori2019look}. In addition, confusion is often described as a temporally evolving process, ranging from subtle early signs, such as slight frowning, to more developed patterns involving forward leaning and hand-to-face behaviors.

\begin{figure*}[t]
    \centering
    \includegraphics[width=0.85\linewidth]{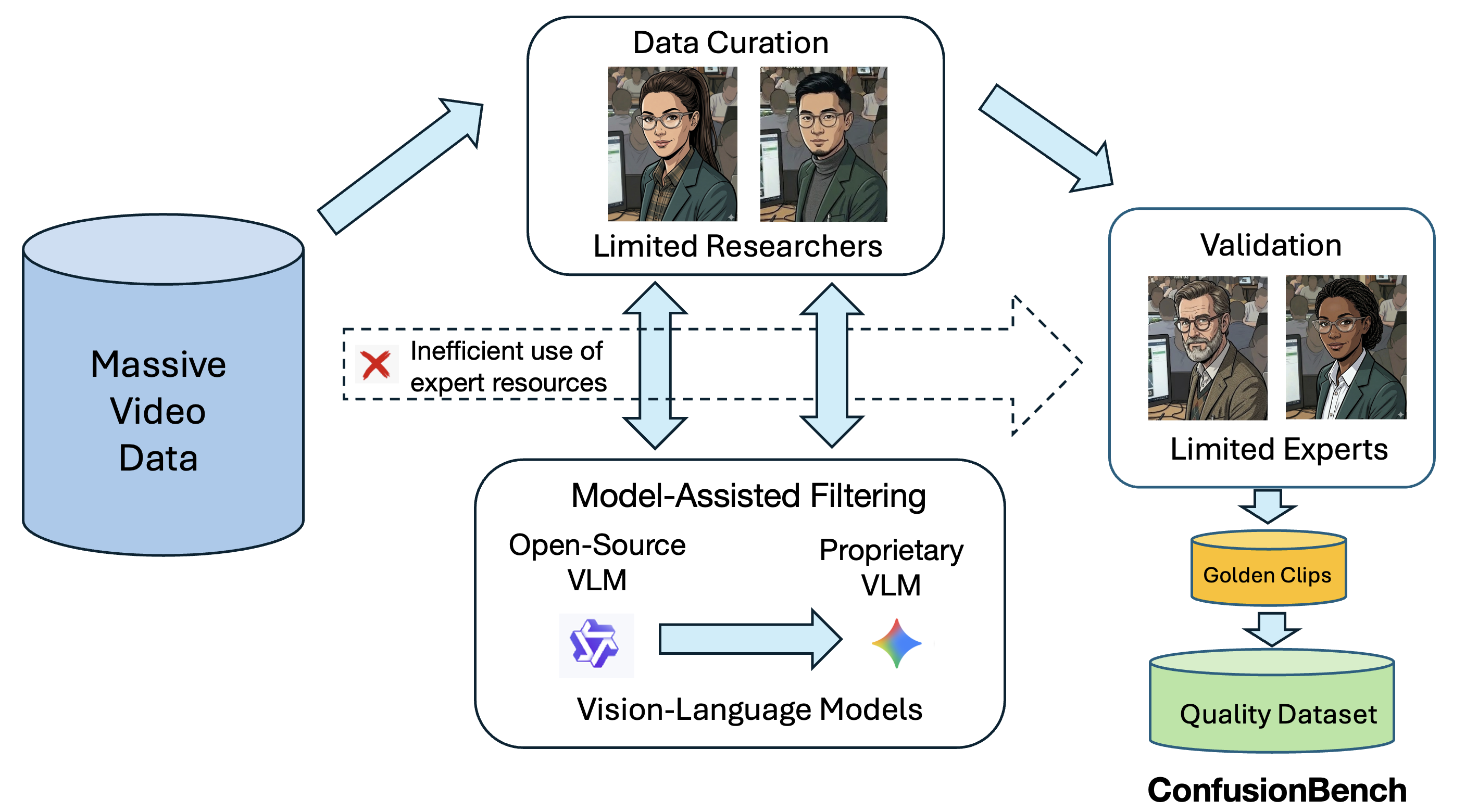}
    \caption{ConfuBench Construction Pipeline}
    \label{fig:overview}
\end{figure*}

\paragraph{DAiSEE Dataset}
DAiSEE \cite{gupta2016daisee} is one of the most widely used video datasets for affective-state understanding in e-learning, containing 9,068 clips from 112 users with annotations for boredom, confusion, engagement, and frustration. 
Although DevEmo \cite{manikowska2023devemo} also provides confusion annotations, it contains only around 50 confusion clips and is therefore much smaller in scale. 
DAiSEE has been widely used for learner-state analysis, including engagement modeling and confusion-related studies \cite{abedi2021improving,khenkar2023deep,komaravalli2023detecting,malekshahi2024general,zheng2024addressing,su2024leveraging,fang2025cross,xu2023automatic}. However, several follow-up studies have noted limitations of DAiSEE, particularly in annotation quality and label distribution \cite{khan2024revisiting,fang2025cross,malekshahi2024general}. In addition, its clip-level annotations may be too coarse for fine-grained confusion analysis, since brief confusion cues can occupy only a small portion of a longer segment \cite{mohamad2019automatic,komaravalli2023detecting}. 
These limitations highlight the need for high-quality, fine-grained, and balanced benchmarks for confusion analysis.

\paragraph{Vision-language models (VLMs)}
VLMs have advanced rapidly in recent years and achieved strong performance across a wide range of multimodal understanding tasks. 
In particular, recent VLMs have shown promising capabilities in video understanding, temporal reasoning, and zero-shot inference, making them attractive tools for evaluating challenging video benchmarks without task-specific training \cite{ren2024timechat,lin2024video,zhang2024long}. 
Beyond direct inference, prior work has also explored using strong multimodal models as teachers to distill knowledge \cite{li2024promptkd,yang2024clip} or generate pseudo-labels for downstream datasets \cite{zhang2024candidate,xing2024vision}. In our setting, we leverage VLMs in two ways. First, we use VLM-based reasoning as an auxiliary tool in the data construction pipeline to help reduce annotation noise during candidate filtering. Second, we evaluate representative open-source and proprietary VLMs on our benchmark to better understand their current ability to recognize subtle confusion cues and perform temporally grounded confusion analysis in educational videos.

\section {ConfusionBench Construction }

To bridge the gap between large-scale candidate data and limited expert resources, we propose a VLM-assisted multi-stage filtering pipeline that removes irrelevant samples and identifies representative clips for expert validation. See Fig.\ref{fig:overview}. Rather than relying directly on the original clip annotations, we progressively refine candidate samples through multiple stages, including clip segmentation, model-assisted filtering, researcher curation, and expert validation. This pipeline allows us to construct a higher-quality set of golden clips and derive temporally grounded long-video annotations. In the following, we describe each stage of the pipeline in detail.

\subsection{Fine-Grained Two-Second Clip Segmentation}

Existing 10-second clip annotations are often too coarse for subtle confusion analysis, since brief confusion cues may occupy only a small portion of a longer segment. 
Prior work on fine-grained emotion recognition suggests using smaller temporal units, and studies of facial expressions indicate that many natural expressions last only a few seconds ~\cite{zhang2022weakly,zhang2020corrnet,schmidt2009comparison,yan2014casme}. Based on these observations, we adopt 2-second clips as a practically reasonable and finer-grained unit for confusion analysis. Concretely, each original 10-second DAiSEE clip is divided into five non-overlapping 2-second clips, which better preserve subtle and temporally brief confusion signals for subsequent filtering and validation.

\subsection{Model-Assisted Filtering}
\begin{figure}[t]
    \centering
    \includegraphics[width=0.9\linewidth]{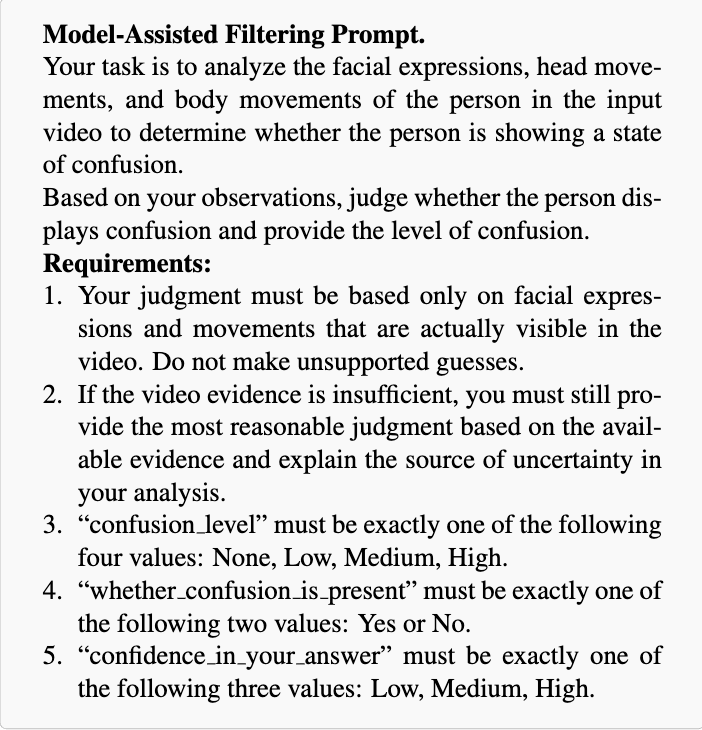}
    \caption{Prompt Design}
    \label{fig:prompt}
\end{figure}

Since our benchmark focuses on confusion analysis and the raw candidate pool is highly imbalanced, we first use a free open-source VLM for coarse filtering. Specifically, Qwen (Qwen3-VL-4B-Instruct) is used to identify 2-second clips likely to contain confusion-related cues. Inference is performed on a local computer with an NVIDIA RTX 3090 GPU. 
We also employ carefully designed prompts to guide model reasoning and enforce a unified output format. The prompt used in this stage is as shown in Fig. \ref{fig:prompt}. 
In this process, the outputs includes four confusion levels: \textit{None}, \textit{Low}, \textit{Medium}, and \textit{High}. To focus on more apparent confusion cases, we treat only \textit{Medium} and \textit{High} predictions as positive. 
Once a 2-second clip is flagged as containing confusion, it is mapped back to its corresponding 10-second source segment, which is then passed to a stronger VLM for further analysis. 
This stage retains 609 10-second video clips and serves as an efficient first-pass filtering step in our multi-stage benchmark construction pipeline.

In the next stage, we employ a proprietary VLM, Google Gemini (Gemini 3 Flash Preview), to further refine candidate analysis. Gemini is applied to each 2-second clip using the same prompt as in the previous stage. As the 609 selected 10-second source segments are each divided into five 2-second clips, this stage processes 3,045 clips in total. 
To improve robustness, we adopt a confidence-aware majority voting strategy over the five inference runs. Each prediction is weighted by its confidence level, with High, Medium, and Low assigned scores of 3, 2, and 1, respectively. The scores are then accumulated across the five runs, and the label with the highest total score is selected as the final prediction. 
Although our final task is binary confusion detection, a more fine-grained prompt encourages better reasoning and more stable predictions. This stage ultimately selects 1,348 2-second clips predicted to contain confusion.

\subsection{Researcher Curation}

After model-assisted filtering, we further refine the candidate clips through researcher curation based on a confusion-focused behavioral protocol derived from prior literature on confusion~\cite{d2014confusion,d2012dynamics,lehman2012confusion,grafsgaard2011predicting,mahmoud2011interpreting}. 
Our protocol considers five criteria: (1) brow lowering or frowning (AU4), eyelid tightening or squinting (AU7), and especially their combination (AU4+AU7) as the most reliable indicators; (2) auxiliary facial cues, including upper lip raising (AU10) and lip pressing or tightening (AU24); (3) smiling as an exclusion cue in most cases; (4) auxiliary hand-to-face behaviors, such as chin touching, forehead pressing, and covering the mouth; and (5) auxiliary body movement and posture cues, such as forward leaning, upright alert posture, and restless movements. 
The goal of this stage is to remove clearly irrelevant samples, reduce redundant neighboring clips, and retain clips with plausible confusion-related cues for expert validation. Two researchers independently review the clips, and only samples with full agreement are retained. In total, 301 clips are selected for expert annotation.

\subsection{Expert Validation}

In the next stage, we submit the 301 curated clips with relatively high confidence to experts in facial expression analysis and human cognition for further validation. Each expert is asked to review the clip and assign one of three decisions: \textit{Yes}, \textit{No}, or \textit{Unsure}, depending on whether the clip exhibits confusion-related cues. \textit{Unsure} indicates that additional information is needed for a more confident judgment. This design allows us to separate clear positive samples from clear negative samples.
Two experts participated in this process, and the agreed-upon validation results are summarized as follows:
224 clips are labeled as \textit{Yes}, 46 clips as \textit{Unsure}, and 31 clips as \textit{No}. The \textit{Yes} clips form the foundation of our final golden clip benchmark and are retained as the core confusion-positive set.

To construct a balanced dataset, we combine the expert-validated \textit{Yes} and \textit{No} samples with an additional set of non-confusion clips of comparable scale. This step is relatively straightforward, as non-confusion samples are conservatively selected from clips that exhibit none of the five confusion-related criteria described above. As a result, we obtain a balanced confusion recognition dataset of 450 clips, including 224 \textit{Yes} samples and 226 \textit{No} samples.

\subsection{Long-Video Confusion Localization}

DAiSEE is originally derived from longer recordings for each participant and segmented into 10-second clips, with clip identifiers following a consistent naming rule that allows the original temporal order to be recovered. In other words, the long videos can be reconstructed from clip IDs. Leveraging this property, we map each expert-validated golden clip back to its corresponding participant and source segment, and reconstruct long videos by concatenating consecutive clips from the same recording. 
We then use the validated golden clips as reference anchors and further refine the start and end boundaries of each confusion event according to the confusion cues and curation principles described in the previous stages. In this way, we obtain higher-quality timestamp-level confusion localization annotations for long-video evaluation. In our benchmark, we construct a collection of ten 5-minute long videos to create a more realistic setting for long-video confusion localization.
\begin{figure}[t]
    \centering
    \includegraphics[width=1.0\linewidth]{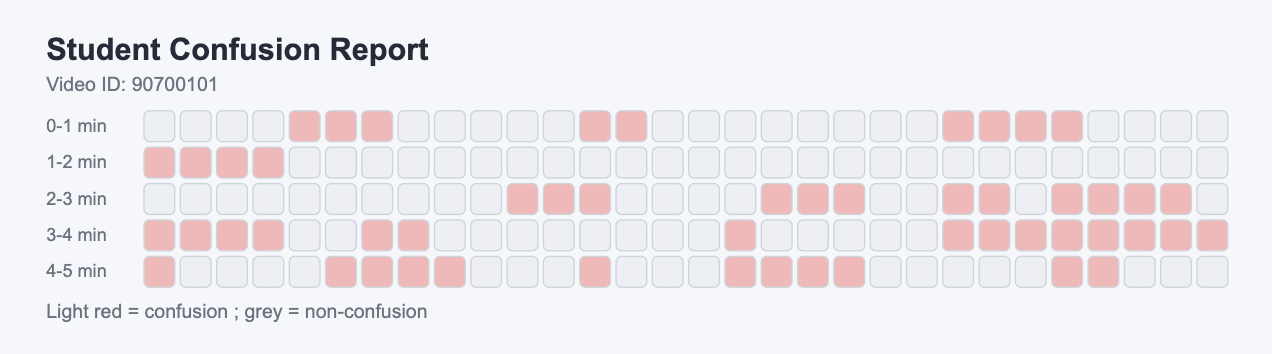}

    \vspace{0.5em}
    \includegraphics[width=1.0\linewidth]
    {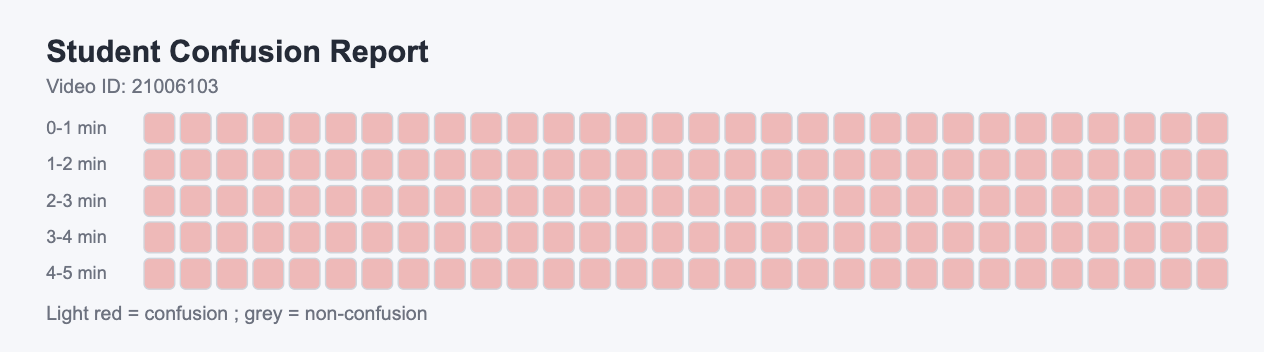}
    \caption{Sample student confusion visualization reports. These reports provide fast, interpretable cues that help educational experts determine whether intervention is needed and adjust instructional plans accordingly.}
    \label{fig:intrograph}
    
\end{figure}

\begin{table}[b]
\centering
\caption{VLM Zero-shot clip-level confusion recognition results on the proposed balanced dataset.}
\label{tab:balance_dataset}
\begin{tabular}{lcccc}
\toprule
\textbf{VLM} & \textbf{Acc} & \textbf{Prec} & \textbf{Rec} & \textbf{F1} \\
\midrule
Qwen & 0.6911 & 0.7815 & 0.5268 & 0.6293\\
Gemini & 0.7978 & 0.7509 & 0.8884 & 0.8139 \\
\bottomrule
\end{tabular}
\end{table}

\begin{table*}[t]
\centering
\caption{Zero-shot long-video confusion localization results using a proprietary VLM (Gemini 3 Flash Preview).}
\label{tab:gemini}
\resizebox{\textwidth}{!}{
\begin{tabular}{lcccc|ccc|ccc|ccc}
\toprule
\textbf{Source} & \textbf{Acc} & \textbf{Prec} & \textbf{Rec} & \textbf{F1} & \textbf{GT\_ev} & \textbf{Pr\_ev} & \textbf{tIoU} & \textbf{P@.1} & \textbf{R@.1} & \textbf{F1@.1} & \textbf{P@.3} & \textbf{R@.3} & \textbf{F1@.3} \\
\midrule
001 & 0.9067 & 0.8684 & 0.7857 & 0.8250 & 4  & 10 & 0.7021 & 0.4000 & 1.0000 & 0.5714 & 0.2000 & 0.5000 & 0.2857 \\
002  & 0.7467 & 0.9111 & 0.5467 & 0.6833 & 11 & 19 & 0.5190 & 0.5263 & 0.9091 & 0.6667 & 0.3684 & 0.6364 & 0.4667 \\
003  & 0.7200 & 0.7273 & 0.9072 & 0.8073 & 17 & 19 & 0.6769 & 0.6842 & 0.7647 & 0.7222 & 0.5263 & 0.5882 & 0.5556 \\
004 & 0.6267 & 0.6116 & 0.8916 & 0.7255 & 17 & 23 & 0.5692 & 0.6087 & 0.8235 & 0.7000 & 0.4783 & 0.6471 & 0.5500 \\
005  & 0.6800 & 0.5926 & 0.9412 & 0.7273 & 16 & 22 & 0.5714 & 0.3636 & 0.5000 & 0.4211 & 0.3182 & 0.4375 & 0.3684 \\
006  & 0.9000 & 0.7903 & 0.9608 & 0.8673 & 11 & 20 & 0.7656 & 0.5000 & 0.9091 & 0.6452 & 0.5000 & 0.9091 & 0.6452 \\
007  & 0.6867 & 0.6327 & 0.8493 & 0.7251 & 11 & 22 & 0.5688 & 0.4091 & 0.8182 & 0.5455 & 0.3636 & 0.7273 & 0.4848 \\
008  & 0.7133 & 0.7407 & 0.3571 & 0.4819 & 7  & 19 & 0.3175 & 0.2632 & 0.7143 & 0.3846 & 0.1053 & 0.2857 & 0.1538 \\
009  & 0.5933 & 0.8727 & 0.4706 & 0.6115 & 6  & 25 & 0.4404 & 0.1600 & 0.6667 & 0.2581 & 0.1200 & 0.5000 & 0.1935 \\
010 & 0.3200 & 1.0000 & 0.3200 & 0.4848 & 1  & 22 & 0.3200 & 0.0000 & 0.0000 & 0.0000 & 0.0000 & 0.0000 & 0.0000 \\
\midrule
\textbf{[Micro-Avg]} & \textbf{0.6893} & \textbf{0.7289} & \textbf{0.6612} & \textbf{0.6934} & \textbf{101} & \textbf{201} & \textbf{0.5307} & \textbf{0.3831} & \textbf{0.7624} & \textbf{0.5099} & \textbf{0.2985} & \textbf{0.5941} & \textbf{0.3974} \\
\textbf{[Macro-Avg]} & \textbf{0.6893} & \textbf{0.7747} & \textbf{0.7030} & \textbf{0.6939} & \textbf{10.1} & \textbf{20.1} & \textbf{0.5451} & \textbf{0.3915} & \textbf{0.7106} & \textbf{0.4915} & \textbf{0.2980} & \textbf{0.5231} & \textbf{0.3704} \\
\bottomrule
\end{tabular}
}
\end{table*}

\begin{table*}[t]
\centering
\caption{Zero-shot long-video confusion localization results using a open source VLM (Qwen3-VL-4B-Instruct).}
\label{tab:qwen}
\resizebox{\textwidth}{!}{
\begin{tabular}{lcccc|ccc|ccc|ccc}
\toprule
\textbf{Source} & \textbf{Acc} & \textbf{Prec} & \textbf{Rec} & \textbf{F1} & \textbf{GT\_ev} & \textbf{Pr\_ev} & \textbf{tIoU} & \textbf{P@.1} & \textbf{R@.1} & \textbf{F1@.1} & \textbf{P@.3} & \textbf{R@.3} & \textbf{F1@.3} \\
\midrule
001 & 0.7267 & 1.0000 & 0.0238 & 0.0465 & 4  & 1  & 0.0238 & 0.0000 & 0.0000 & 0.0000 & 0.0000 & 0.0000 & 0.0000 \\
002  & 0.4933 & 0.0000 & 0.0000 & 0.0000 & 11 & 1  & 0.0000 & 0.0000 & 0.0000 & 0.0000 & 0.0000 & 0.0000 & 0.0000 \\
003  & 0.6067 & 0.7317 & 0.6186 & 0.6704 & 17 & 22 & 0.5042 & 0.5455 & 0.7059 & 0.6154 & 0.4091 & 0.5294 & 0.4615 \\
004 & 0.6333 & 0.6346 & 0.7952 & 0.7059 & 17 & 19 & 0.5455 & 0.6842 & 0.7647 & 0.7222 & 0.5263 & 0.5882 & 0.5556 \\
005  & 0.6733 & 0.6462 & 0.6176 & 0.6316 & 16 & 23 & 0.4615 & 0.5217 & 0.7500 & 0.6154 & 0.4348 & 0.6250 & 0.5128 \\
006  & 0.8067 & 0.7750 & 0.6078 & 0.6813 & 11 & 15 & 0.5167 & 0.6000 & 0.8182 & 0.6923 & 0.4667 & 0.6364 & 0.5385 \\
007  & 0.5933 & 0.6034 & 0.4795 & 0.5344 & 11 & 15 & 0.3646 & 0.4000 & 0.5455 & 0.4615 & 0.2000 & 0.2727 & 0.2308 \\
008 & 0.6400 & 0.7500 & 0.0536 & 0.1000 & 7  & 4  & 0.0526 & 0.0000 & 0.0000 & 0.0000 & 0.0000 & 0.0000 & 0.0000 \\
009  & 0.3667 & 0.6522 & 0.1471 & 0.2400 & 6  & 9  & 0.1364 & 0.2222 & 0.3333 & 0.2667 & 0.0000 & 0.0000 & 0.0000 \\
010 & 0.1133 & 1.0000 & 0.1133 & 0.2036 & 1  & 12 & 0.1133 & 0.0000 & 0.0000 & 0.0000 & 0.0000 & 0.0000 & 0.0000 \\
\midrule
\textbf{[Micro-Avg]} & \textbf{0.5653} & \textbf{0.6835} & \textbf{0.3388} & \textbf{0.4530} & \textbf{101} & \textbf{121} & \textbf{0.2928} & \textbf{0.4463} & \textbf{0.5347} & \textbf{0.4865} & \textbf{0.3223} & \textbf{0.3861} & \textbf{0.3514} \\
\textbf{[Macro-Avg]} & \textbf{0.5653} & \textbf{0.6793} & \textbf{0.3456} & \textbf{0.3814} & \textbf{10.1} & \textbf{12.1} & \textbf{0.2719} & \textbf{0.2974} & \textbf{0.3918} & \textbf{0.3374} & \textbf{0.2037} & \textbf{0.2652} & \textbf{0.2299} \\
\bottomrule
\end{tabular}
}

\end{table*}

\section{Experiments}

\begin{figure*}[htbp]
    \centering
    \includegraphics[width=1\linewidth]{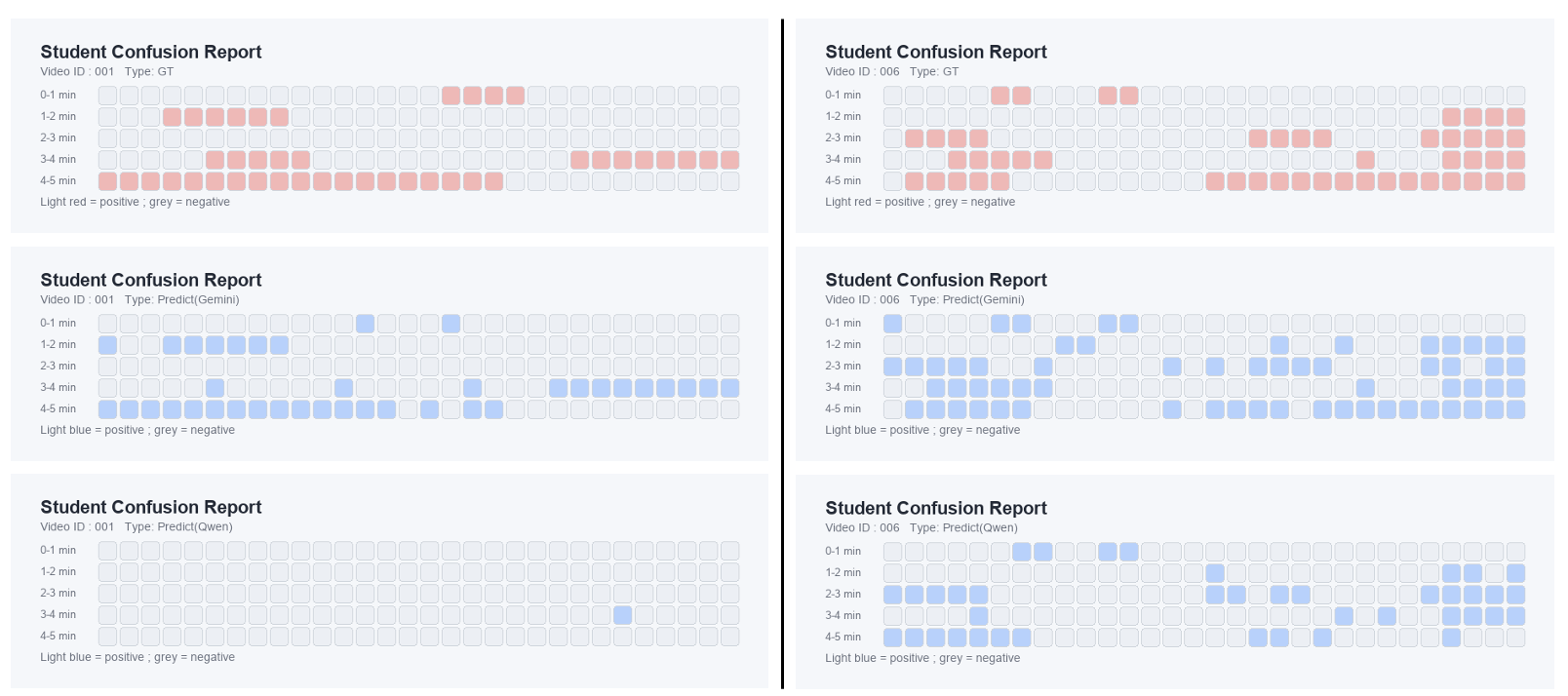}
    \caption{ \textbf{Comparison of Student Confusion Reports.} The first row in red presents the ground-truth annotations. The second row in blue shows the predictions from Gemini (Gemini 3 Flash Preview), a proprietary multimodal model optimized for high-throughput inference. The third row in blue shows the predictions from Qwen (Qwen3-VL-4B-Instruct), a lightweight local multimodal model with strong video understanding capabilities.}
    \label{fig:compare}
\end{figure*}

Given the rapid progress of vision-language models (VLMs) and their strong multimodal reasoning ability, their capability for confusion understanding remains underexplored. We therefore report zero-shot baselines on the proposed benchmark. Specifically, we evaluate Qwen3-VL-4B-Instruct (\textit{Qwen}) and Gemini 3 Flash Preview (\textit{Gemini}) without any task-specific training or fine-tuning, to assess how well current VLMs can directly recognize confusion-related cues.

For the confusion recognition task, we evaluate model performance using standard classification metrics, including Accuracy (Acc) , Precision (Prec), Recall(Rec), and F1 score. These metrics provide complementary views of model behavior, allowing us to measure overall correctness as well as the model’s ability to reliably identify confusion samples that are relatively salient to human observers.

For long-video confusion localization, we evaluate both event detection and temporal boundary quality. Consecutive confusion-positive segments are grouped into predicted events (Pr\_ev) and compared with ground-truth events (GT\_ev). Temporal overlap is measured by (tIoU), defined as the ratio between the overlap duration and the union duration of a predicted event and a ground-truth event. We further report event-level Precision, Recall, and F1 under tIoU thresholds of 0.1 and 0.3, corresponding to relatively loose and moderately strict matching criteria. (P@tIoU) measures the proportion of predicted events that correctly match a ground-truth confusion event under the specified overlap threshold, while {R@tIoU} measures the proportion of ground-truth confusion events that are successfully detected by the model. (F1@tIoU) provides the harmonic mean of Precision and Recall.
Finally, we report both micro-average (Micro-Avg) and macro-average (Macro-Avg) metrics for overall aggregation.

The confusion recognition results in Table~\ref{tab:balance_dataset} show that \textit{Gemini} outperforms \textit{Qwen} by a large margin in terms of F1 score (0.8139 vs. 0.6293). This gap is mainly attributable to Qwen’s relatively low recall (0.5268), which suggests that many clips containing confusion are misclassified as non-confusion. These results indicate that \textit{Qwen} is more conservative in recognizing confusion and lacks sufficient sensitivity to subtle confusion-related cues.

The confusion video localization results in Table~\ref{tab:gemini} suggest that \textit{Gemini} achieves reasonably strong segment-level confusion recognition, with a Micro-Avg F1 of 0.6934. At the event level, the model attains a Micro-Avg tIoU of 0.5307, indicating moderate temporal alignment for matched confusion events. However, the gap between F1@0.1 (0.5099) and F1@0.3 (0.3974) shows that, although Gemini can often roughly identify the presence of confusion events, accurately localizing their temporal boundaries remains challenging. In addition, the relatively high Recall@0.1 (0.7624) compared with Precision@0.1 (0.3831) suggests that \textit{Gemini} is more sensitive than precise in long-video confusion detection, tending to cover more true events at the cost of additional false positives.

The long-video confusion localization results reported in Table~\ref{tab:qwen} show that \textit{Qwen} performs clearly worse than \textit{Gemini}. Its Micro-Avg recall is low (0.3388) despite a moderate precision (0.6835), suggesting that the model is conservative in predicting confusion and misses many true confusion segments. This is also reflected in the event-level results, where \textit{Qwen} achieves only 0.4865 F1@0.1 and 0.3514 F1@0.3, indicating limited ability in both event discovery and temporal localization. In addition, the relatively low Micro-Avg tIoU (0.2928) suggests weak temporal alignment between predicted and ground-truth confusion events. This may partly reflect the limited capacity of the 4B model. Overall, these results suggest that this version of \textit{Qwen} remains limited in both sensitivity and temporal grounding for long-video confusion localization.

The student confusion visualization report in Figure~\ref{fig:compare} further validates the above discussion. \textit{Gemini} generally produces more accurate confusion predictions than the open-source \textit{Qwen} for both clearly confused and clearly non-confused states. In particular, \textit{Gemini} recovers obvious confusion segments more effectively while maintaining more reasonable coverage of negative regions, whereas \textit{Qwen} misses many true confusion intervals and appears overly conservative. 
This difference is especially evident in transitional or ambiguous segments, where \textit{Gemini} more often labels subtle indicators, such as hand-on-chin or other emerging confusion-related behaviors, as confusion, while \textit{Qwen} tends to refrain from assigning confusion labels to these intermediate states, leading to substantial missed detections, as illustrated in the left example of Figure~\ref{fig:compare}.
Overall, these qualitative results are consistent with the quantitative findings: \textit{Gemini} shows stronger sensitivity and more complete temporal coverage, whereas \textit{Qwen} remains limited by conservative predictions and weaker temporal grounding.

\section{Conclusion}

In this work, we identify key limitations of existing confusion datasets, including noisy labels, coarse temporal annotations, and insufficient expert validation. To address these issues, we propose a VLM-assisted multi-stage benchmark construction pipeline that balances expert effort, model cost, and data quality. Based on this pipeline, we construct an expert-validated balanced confusion recognition dataset, a collection of long videos with fine-grained temporal annotations for confusion localization, and a confusion report visualization design.
We further benchmark representative VLMs in the zero-shot setting on both confusion recognition and localization tasks, providing a clearer picture of current model capabilities in recognizing confusion. This marks an important step toward more effective automated interactive systems.

In future work, we plan to expand the video localization benchmark, evaluate a wider range of models, and develop improved confusion localization frameworks for more reliable student confusion reports.

\section{Limitations}

Our current recognition dataset contains 450 video clips, including 224 expert-validated golden confusion clips and 226 non-confusion clips, while the video localization benchmark includes 10 long videos of approximately 5 minutes each. Although these datasets provide an initial testbed for confusion recognition and localization, their scale remains limited and could be improved by expanding both the golden clip set and the long-video benchmark.

In addition, our current formulation focuses on whether confusion is present, without distinguishing among different stages or intensities of confusion. In practice, confusion may manifest as early, moderate, or strong confusion. In this work, all such cases are treated as a single category. Developing a more fine-grained evaluation of confusion, in collaboration with facial expression and educational experts, remains an important direction for future work.

\section{Ethics Consideration}

This work does not involve new human-subject data collection. We use data curated from the publicly available DAiSEE dataset, which has been released since 2016. According to the dataset documentation, participants provided signed consent for their videos to be shared with the research community. We follow the dataset usage agreement and use the data only for academic research.
For vision-language model analysis, we use both local and cloud-based models. We run Qwen locally on our computer for data analysis. We access Gemini through Google Cloud Vertex AI. According to Google, customer data is not used to train its AI models, and API communications are encrypted in transit using TLS. 

Overall, we consider the ethical risk of this work to be limited, since it relies on previously released research data and does not involve new participant recruitment. Nevertheless, possible misinterpretation of students’ affective states, especially under limited or variable video quality, remains an important consideration. Accordingly, the proposed benchmark is intended for research use only, and model predictions should be treated as supportive signals rather than as a basis for high-stakes educational decisions.

\newpage

{
    \small
    \bibliographystyle{ieeenat_fullname}
    \bibliography{main}

\begin{thebibliography}{47}
\providecommand{\natexlab}[1]{#1}
\providecommand{\url}[1]{\texttt{#1}}
\expandafter\ifx\csname urlstyle\endcsname\relax
  \providecommand{\doi}[1]{doi: #1}\else
  \providecommand{\doi}{doi: \begingroup \urlstyle{rm}\Url}\fi

\bibitem[Abedi and Khan(2021)]{abedi2021improving}
Ali Abedi and Shehroz~S Khan.
\newblock Improving state-of-the-art in detecting student engagement with resnet and tcn hybrid network.
\newblock In \emph{2021 18th Conference on Robots and Vision (CRV)}, pages 151--157. IEEE, 2021.

\bibitem[Cowen et~al.(2019)Cowen, Sauter, Tracy, and Keltner]{cowen2019mapping}
Alan Cowen, Disa Sauter, Jessica~L Tracy, and Dacher Keltner.
\newblock Mapping the passions: Toward a high-dimensional taxonomy of emotional experience and expression.
\newblock \emph{Psychological Science in the Public Interest}, 20\penalty0 (1):\penalty0 69--90, 2019.

\bibitem[D'Mello et~al.(2007)D'Mello, Chipman, and Graesser]{d2007posture}
Sidney~S D'Mello, Patrick Chipman, and Art Graesser.
\newblock Posture as a predictor of learner's affective engagement.
\newblock In \emph{Proceedings of the Annual Meeting of the Cognitive Science Society}, 2007.

\bibitem[Dong et~al.(2024{\natexlab{a}})Dong, Chaudhary, Xu, Wang, Lary, and Nwogu]{dong2024signavatar}
Lu Dong, Lipisha Chaudhary, Fei Xu, Xiao Wang, Mason Lary, and Ifeoma Nwogu.
\newblock Signavatar: Sign language 3d motion reconstruction and generation.
\newblock In \emph{2024 IEEE 18th International Conference on Automatic Face and Gesture Recognition (FG)}, pages 1--10. IEEE, 2024{\natexlab{a}}.

\bibitem[Dong et~al.(2024{\natexlab{b}})Dong, Wang, and Nwogu]{dong2024word}
Lu Dong, Xiao Wang, and Ifeoma Nwogu.
\newblock Word-conditioned 3d american sign language motion generation.
\newblock In \emph{Findings of the Association for Computational Linguistics: EMNLP 2024}, pages 9993--9999, 2024{\natexlab{b}}.

\bibitem[Dong et~al.(2024{\natexlab{c}})Dong, Wang, Setlur, Govindaraju, and Nwogu]{dong2024ig3d}
Lu Dong, Xiao Wang, Srirangaraj Setlur, Venu Govindaraju, and Ifeoma Nwogu.
\newblock Ig3d: Integrating 3d face representations in facial expression inference.
\newblock In \emph{European Conference on Computer Vision}, pages 404--421. Springer, 2024{\natexlab{c}}.

\bibitem[D’Mello and Graesser(2012)]{d2012dynamics}
Sidney D’Mello and Art Graesser.
\newblock Dynamics of affective states during complex learning.
\newblock \emph{Learning and Instruction}, 22\penalty0 (2):\penalty0 145--157, 2012.

\bibitem[D’Mello et~al.(2009)D’Mello, Craig, Fike, and Graesser]{d2009responding}
Sidney D’Mello, Scotty Craig, Karl Fike, and Arthur Graesser.
\newblock Responding to learners’ cognitive-affective states with supportive and shakeup dialogues.
\newblock In \emph{International Conference on Human-Computer Interaction}, pages 595--604. Springer, 2009.

\bibitem[D’Mello and Graesser(2014)]{d2014confusion}
Sidney~K D’Mello and Arthur~C Graesser.
\newblock Confusion.
\newblock In \emph{International handbook of emotions in education}, pages 289--310. Routledge, 2014.

\bibitem[Fang et~al.(2025)Fang, Huang, and Ogan]{fang2025cross}
Yu Fang, Shihong Huang, and Amy Ogan.
\newblock A cross-cultural confusion model for detecting and evaluating students’ confusion in a large classroom.
\newblock In \emph{Proceedings of the 15th International Learning Analytics and Knowledge Conference}, pages 473--483, 2025.

\bibitem[Frank and Shaw(2016)]{frank2016evolution}
Mark~G. Frank and Allison~Z. Shaw.
\newblock Evolution and nonverbal communication.
\newblock In \emph{APA Handbook of Nonverbal Communication}, pages 45--76. American Psychological Association, Washington, DC, 2016.

\bibitem[Grafsgaard et~al.(2011)Grafsgaard, Boyer, and Lester]{grafsgaard2011predicting}
Joseph~F Grafsgaard, Kristy~Elizabeth Boyer, and James~C Lester.
\newblock Predicting facial indicators of confusion with hidden markov models.
\newblock In \emph{International Conference on Affective computing and intelligent interaction}, pages 97--106. Springer, 2011.

\bibitem[Gupta et~al.(2016)Gupta, D'Cunha, Awasthi, and Balasubramanian]{gupta2016daisee}
Abhay Gupta, Arjun D'Cunha, Kamal Awasthi, and Vineeth Balasubramanian.
\newblock Daisee: Towards user engagement recognition in the wild.
\newblock \emph{arXiv preprint arXiv:1609.01885}, 2016.

\bibitem[Karg et~al.(2013)Karg, Samadani, Gorbet, K{\"u}hnlenz, Hoey, and Kuli{\'c}]{karg2013body}
Michelle Karg, Ali-Akbar Samadani, Rob Gorbet, Kolja K{\"u}hnlenz, Jesse Hoey, and Dana Kuli{\'c}.
\newblock Body movements for affective expression: A survey of automatic recognition and generation.
\newblock \emph{IEEE Transactions on Affective Computing}, 4\penalty0 (4):\penalty0 341--359, 2013.

\bibitem[Khan and Safa(2024)]{khan2024revisiting}
Shehroz Khan and Sadaf Safa.
\newblock Revisiting annotations in online student engagement.
\newblock In \emph{Proceedings of the 2024 10th International Conference on Computing and Data Engineering}, pages 111--117, 2024.

\bibitem[Khenkar et~al.(2023)Khenkar, Jarraya, Allinjawi, Alkhuraiji, Abuzinadah, and Kateb]{khenkar2023deep}
Shoroog~Ghazee Khenkar, Salma~Kammoun Jarraya, Arwa Allinjawi, Samar Alkhuraiji, Nihal Abuzinadah, and Faris~A Kateb.
\newblock Deep analysis of student body activities to detect engagement state in e-learning sessions.
\newblock \emph{Applied Sciences}, 13\penalty0 (4):\penalty0 2591, 2023.

\bibitem[Komaravalli and Janet(2023)]{komaravalli2023detecting}
Purushottama~Rao Komaravalli and B Janet.
\newblock Detecting academic affective states of learners in online learning environments using deep transfer learning.
\newblock \emph{Scalable Computing: Practice and Experience}, 24\penalty0 (4):\penalty0 957--970, 2023.

\bibitem[Lehman et~al.(2012)Lehman, D'Mello, and Graesser]{lehman2012confusion}
Blair Lehman, Sidney D'Mello, and Art Graesser.
\newblock Confusion and complex learning during interactions with computer learning environments.
\newblock \emph{The Internet and Higher Education}, 15\penalty0 (3):\penalty0 184--194, 2012.

\bibitem[Li et~al.(2024)Li, Li, Fu, Zhang, Wang, Chen, and Yang]{li2024promptkd}
Zheng Li, Xiang Li, Xinyi Fu, Xin Zhang, Weiqiang Wang, Shuo Chen, and Jian Yang.
\newblock Promptkd: Unsupervised prompt distillation for vision-language models.
\newblock In \emph{Proceedings of the IEEE/CVF Conference on Computer Vision and Pattern Recognition}, pages 26617--26626, 2024.

\bibitem[Lin et~al.(2024)Lin, Ye, Zhu, Cui, Ning, Jin, and Yuan]{lin2024video}
Bin Lin, Yang Ye, Bin Zhu, Jiaxi Cui, Munan Ning, Peng Jin, and Li Yuan.
\newblock Video-llava: Learning united visual representation by alignment before projection.
\newblock In \emph{Proceedings of the 2024 conference on empirical methods in natural language processing}, pages 5971--5984, 2024.

\bibitem[Mahmoud and Robinson(2011)]{mahmoud2011interpreting}
Marwa Mahmoud and Peter Robinson.
\newblock Interpreting hand-over-face gestures.
\newblock In \emph{International Conference on Affective Computing and Intelligent Interaction}, pages 248--255. Springer, 2011.

\bibitem[Malekshahi et~al.(2024)Malekshahi, Kheyridoost, and Fatemi]{malekshahi2024general}
Somayeh Malekshahi, Javad~M Kheyridoost, and Omid Fatemi.
\newblock A general model for detecting learner engagement: implementation and evaluation.
\newblock \emph{arXiv preprint arXiv:2405.04251}, 2024.

\bibitem[Manikowska et~al.(2023)Manikowska, Sadowski, Sowinski, and Wrobel]{manikowska2023devemo}
Michalina Manikowska, Damian Sadowski, Adam Sowinski, and Michal~R Wrobel.
\newblock Devemo—software developers’ facial expression dataset.
\newblock \emph{Applied Sciences}, 13\penalty0 (6):\penalty0 3839, 2023.

\bibitem[Matsumoto et~al.(2012)Matsumoto, Frank, and Hwang]{matsumoto2012nonverbal}
David Matsumoto, Mark~G Frank, and Hyi~Sung Hwang.
\newblock \emph{Nonverbal communication: Science and applications}.
\newblock Sage Publications, 2012.

\bibitem[Mohamad~Nezami et~al.(2019)Mohamad~Nezami, Dras, Hamey, Richards, Wan, and Paris]{mohamad2019automatic}
Omid Mohamad~Nezami, Mark Dras, Len Hamey, Deborah Richards, Stephen Wan, and C{\'e}cile Paris.
\newblock Automatic recognition of student engagement using deep learning and facial expression.
\newblock In \emph{Joint european conference on machine learning and knowledge discovery in databases}, pages 273--289. Springer, 2019.

\bibitem[Mori and Pell(2019)]{mori2019look}
Yondu Mori and Marc~D Pell.
\newblock The look of (un) confidence: visual markers for inferring speaker confidence in speech.
\newblock \emph{Frontiers in Communication}, 4:\penalty0 487004, 2019.

\bibitem[Myers(2021)]{myers2021automatic}
Mark~H Myers.
\newblock Automatic detection of a student’s affective states for intelligent teaching systems.
\newblock \emph{Brain Sciences}, 11\penalty0 (3):\penalty0 331, 2021.

\bibitem[Prince et~al.(2015)Prince, Martin, Messinger, and Allen]{prince2015facial}
Emily~B Prince, Katherine~B Martin, Daniel~S Messinger, and M Allen.
\newblock Facial action coding system.
\newblock \emph{Environmental psychology \& nonverbal behavior}, 1, 2015.

\bibitem[Ren et~al.(2024)Ren, Yao, Li, Sun, and Hou]{ren2024timechat}
Shuhuai Ren, Linli Yao, Shicheng Li, Xu Sun, and Lu Hou.
\newblock Timechat: A time-sensitive multimodal large language model for long video understanding.
\newblock In \emph{Proceedings of the IEEE/CVF Conference on Computer Vision and Pattern Recognition}, pages 14313--14323, 2024.

\bibitem[Rozin and Cohen(2003)]{rozin2003high}
Paul Rozin and Adam~B Cohen.
\newblock High frequency of facial expressions corresponding to confusion, concentration, and worry in an analysis of naturally occurring facial expressions of americans.
\newblock \emph{Emotion}, 3\penalty0 (1):\penalty0 68, 2003.

\bibitem[Schmidt et~al.(2009)Schmidt, Bhattacharya, and Denlinger]{schmidt2009comparison}
Karen~L Schmidt, Sharika Bhattacharya, and Rachel Denlinger.
\newblock Comparison of deliberate and spontaneous facial movement in smiles and eyebrow raises.
\newblock \emph{Journal of nonverbal behavior}, 33\penalty0 (1):\penalty0 35--45, 2009.

\bibitem[Shan et~al.(2024)Shan, Dong, Han, Yao, Liu, Nwogu, Qi, and Hill]{shan2024towards}
Mengyi Shan, Lu Dong, Yutao Han, Yuan Yao, Tao Liu, Ifeoma Nwogu, Guo-Jun Qi, and Mitch Hill.
\newblock Towards open domain text-driven synthesis of multi-person motions.
\newblock In \emph{European Conference on Computer Vision}, pages 67--86. Springer, 2024.

\bibitem[Su et~al.(2024)Su, He, and Luo]{su2024leveraging}
Rui Su, Lang He, and Mengnan Luo.
\newblock Leveraging part-and-sensitive attention network and transformer for learner engagement detection.
\newblock \emph{Alexandria Engineering Journal}, 107:\penalty0 198--204, 2024.

\bibitem[Wang et~al.(2025)Wang, Dong, Rangasrinivasan, Nwogu, Setlur, and Govindaraju]{wang2025automisty}
Xiao Wang, Lu Dong, Sahana Rangasrinivasan, Ifeoma Nwogu, Srirangaraj Setlur, and Venugopal Govindaraju.
\newblock Automisty: a multi-agent llm framework for automated code generation in the misty social robot.
\newblock In \emph{2025 IEEE/RSJ International Conference on Intelligent Robots and Systems (IROS)}, pages 9194--9201. IEEE, 2025.

\bibitem[Wang et~al.(2026)Wang, Dong, Sun, Nwogu, Setlur, and Govindaraju]{wang2026mistypilot}
Xiao Wang, Lu Dong, Jingchen Sun, Ifeoma Nwogu, Srirangaraj Setlur, and Venu Govindaraju.
\newblock Mistypilot: An agentic fast-slow thinking llm framework for misty social robots.
\newblock \emph{arXiv preprint arXiv:2603.03640}, 2026.

\bibitem[Xing et~al.(2024)Xing, Xiong, Stylianou, Sastry, Gong, and Jacobs]{xing2024vision}
Xin Xing, Zhexiao Xiong, Abby Stylianou, Srikumar Sastry, Liyu Gong, and Nathan Jacobs.
\newblock Vision-language pseudo-labels for single-positive multi-label learning.
\newblock In \emph{Proceedings of the IEEE/CVF conference on computer vision and pattern recognition}, pages 7799--7808, 2024.

\bibitem[Xu et~al.(2023)Xu, Zheng, Li, and Li]{xu2023automatic}
Yaping Xu, Yaqian Zheng, Keru Li, and Yanyan Li.
\newblock Automatic recognition and analysis of academic emotions based on facial expressions during online learning environments.
\newblock In \emph{Proceedings of the 15th International Conference on Education Technology and Computers}, pages 328--334, 2023.

\bibitem[Yan et~al.(2014)Yan, Li, Wang, Zhao, Liu, Chen, and Fu]{yan2014casme}
Wen-Jing Yan, Xiaobai Li, Su-Jing Wang, Guoying Zhao, Yong-Jin Liu, Yu-Hsin Chen, and Xiaolan Fu.
\newblock Casme ii: An improved spontaneous micro-expression database and the baseline evaluation.
\newblock \emph{PloS one}, 9\penalty0 (1):\penalty0 e86041, 2014.

\bibitem[Yang et~al.(2024)Yang, An, Huang, Bi, Yu, Yang, Diao, and Xu]{yang2024clip}
Chuanguang Yang, Zhulin An, Libo Huang, Junyu Bi, Xinqiang Yu, Han Yang, Boyu Diao, and Yongjun Xu.
\newblock Clip-kd: An empirical study of clip model distillation.
\newblock In \emph{Proceedings of the IEEE/CVF Conference on Computer Vision and Pattern Recognition}, pages 15952--15962, 2024.

\bibitem[Yu et~al.(2025)Yu, Androsov, and Yan]{yu2025exploring}
Shuzhen Yu, Alexey Androsov, and Hanbing Yan.
\newblock Exploring the prospects of multimodal large language models for automated emotion recognition in education: Insights from gemini.
\newblock \emph{Computers \& Education}, 232:\penalty0 105307, 2025.

\bibitem[Zeng et~al.(2017)Zeng, Chaturvedi, and Bhat]{zeng2017learner}
Ziheng Zeng, Snigdha Chaturvedi, and Suma Bhat.
\newblock Learner affect through the looking glass: Characterization and detection of confusion in online courses.
\newblock \emph{International Educational Data Mining Society}, 2017.

\bibitem[Zhai et~al.(2023)Zhai, Huang, Luan, Dong, Nwogu, Lyu, Doermann, and Yuan]{zhai2023language}
Yuanhao Zhai, Mingzhen Huang, Tianyu Luan, Lu Dong, Ifeoma Nwogu, Siwei Lyu, David Doermann, and Junsong Yuan.
\newblock Language-guided human motion synthesis with atomic actions.
\newblock In \emph{Proceedings of the 31st ACM International Conference on Multimedia}, pages 5262--5271, 2023.

\bibitem[Zhang et~al.(2024{\natexlab{a}})Zhang, Wei, Liu, and Feng]{zhang2024candidate}
Jiahan Zhang, Qi Wei, Feng Liu, and Lei Feng.
\newblock Candidate pseudolabel learning: Enhancing vision-language models by prompt tuning with unlabeled data.
\newblock \emph{arXiv preprint arXiv:2406.10502}, 2024{\natexlab{a}}.

\bibitem[Zhang et~al.(2024{\natexlab{b}})Zhang, Zhang, Li, Zeng, Yang, Zhang, Wang, Tan, Li, and Liu]{zhang2024long}
Peiyuan Zhang, Kaichen Zhang, Bo Li, Guangtao Zeng, Jingkang Yang, Yuanhan Zhang, Ziyue Wang, Haoran Tan, Chunyuan Li, and Ziwei Liu.
\newblock Long context transfer from language to vision.
\newblock \emph{arXiv preprint arXiv:2406.16852}, 2024{\natexlab{b}}.

\bibitem[Zhang et~al.(2020)Zhang, El~Ali, Wang, Hanjalic, and Cesar]{zhang2020corrnet}
Tianyi Zhang, Abdallah El~Ali, Chen Wang, Alan Hanjalic, and Pablo Cesar.
\newblock Corrnet: Fine-grained emotion recognition for video watching using wearable physiological sensors.
\newblock \emph{Sensors}, 21\penalty0 (1):\penalty0 52, 2020.

\bibitem[Zhang et~al.(2022)Zhang, El~Ali, Wang, Hanjalic, and Cesar]{zhang2022weakly}
Tianyi Zhang, Abdallah El~Ali, Chen Wang, Alan Hanjalic, and Pablo Cesar.
\newblock Weakly-supervised learning for fine-grained emotion recognition using physiological signals.
\newblock \emph{IEEE Transactions on Affective Computing}, 14\penalty0 (3):\penalty0 2304--2322, 2022.

\bibitem[Zheng et~al.(2024)Zheng, Hasegawa, Gu, and Ota]{zheng2024addressing}
Xianwen Zheng, Shinobu Hasegawa, Wen Gu, and Koichi Ota.
\newblock Addressing class imbalances in video time-series data for estimation of learner engagement:“over sampling with skipped moving average”.
\newblock \emph{Education Sciences}, 14\penalty0 (6):\penalty0 556, 2024.

\end{thebibliography}
}

\end{document}